\documentclass[accepted]{uai2023} %

\usepackage[british]{babel}
\usepackage{csquotes}
 
\usepackage[sorting=none, style=authoryear-comp, giveninits=true, backend=biber,maxbibnames=10,maxcitenames=1,uniquelist=minyear,
doi=false,
isbn=false,
url=false,eprint=true]{biblatex}
\AtEveryBibitem{%
  \clearfield{note}%
  \clearfield{pages}
  \clearfield{language}
  \clearfield{number}
  \clearfield{volume}
  \clearfield{url}
  \clearfield{urlyear}
  \clearfield{urlmonth}
}

\DeclareNameAlias{sortname}{last-first}

\renewbibmacro{in:}{}

\usepackage{mathtools} %
\usepackage{amsfonts}
\usepackage{booktabs} %
\usepackage{tikz} %
\usepackage[percent]{overpic}

\newcommand{\mengensymbol}[1]{\ensuremath{\mathbb{#1}}}
\newcommand{\R}{\mengensymbol{R}}
\newcommand{\e}{\operatorname{e}}
\newcommand{\E}{\mathbb{E}}
\newcommand{\cov}{\text{cov}}

\title{The Past Does Matter: Correlation of Subsequent States in \\Trajectory Predictions of Gaussian Process Models}

\author[1]{\href{mailto:<steffen@robots.ox.ac.uk>?Subject=Your UAI 2023 paper}{Steffen~Ridderbusch}{}}
\author[2]{Sina~Ober-Bl\"obaum}
\author[1]{Paul~Goulart}
\affil[1]{%
    Control Group, Dept. of Engineering Science\\
    University of Oxford\\
    Oxford, UK
}
\affil[2]{%
    Numerical Mathematics and Control\\
    University of Paderborn\\
    Paderborn, Germany
}
  
\begin{document}
\maketitle

\begin{abstract}
  Computing the distribution of trajectories from a Gaussian Process model of a dynamical system is an important challenge in utilizing such models.
  Motivated by the computational cost of sampling-based approaches, we consider approximations of the model's output and trajectory distribution.
  We show that previous work on uncertainty propagation, focussed on discrete state-space models, incorrectly included an independence assumption between subsequent states of the predicted trajectories.
  Expanding these ideas to continuous ordinary differential equation models, we illustrate the implications of this assumption and propose a novel piecewise linear approximation of Gaussian Processes to mitigate them. 
\end{abstract}

\section{Introduction}

The context of this work is the combination of dynamical systems theory, with its history of widespread applications in science and engineering, and Gaussian Process (GP) models. To utilize these models, one must be able to make predictions for future trajectories, ideally in a way that incorporates the uncertainty of the model. These predictions should also be cheap to compute and qualitatively accurate.

The computation of trajectories depends on the model class. One example of \emph{continuous} models, our main focus, are ordinary differential equation (ODE) models of the form
\begin{equation}
  \dot{x} = f(x).
  \label{eq:baseODE}
\end{equation}
We can represent $f$ as a GP, which means that we learn the \emph{vector field} of a dynamical system, as seen in \cite{ridderbuschLearningODEModels2021} and \cite{heinonenLearningUnknownODE2018}. 
This approach is conceptually similar to \emph{UniversalODEs}, which use a Neural Network instead of a GP [\cite{rackauckasUniversalDifferentialEquations2020}]. 

We note explicitly that this setting does not result in a stochastic differential equation or random dynamical system, where \emph{aleatory} uncertainty arises from inherent stochasticity [\cite{banksUncertaintyPropagationQuantification2012}]. 
Instead, a GP represents a distribution over a function space, conditioned on available pairs of input-output data, which we assume contains the true underlying deterministic function. 
This means that GP models capture \textit{epistemic} uncertainty, as the uncertainty about the true model decreases with additional data.

The state $x$ of a continuous model at time $t$ is given by the \emph{flow} $ \varphi^t_f (x_0)$, which is the solution of the ODE \eqref{eq:baseODE} for the initial value $x_0$. 
However, the flow map is generally not available analytically and instead must be computed approximately at discrete times via numerical integration. 
The simplest such method is the explicit Euler's method
\begin{equation}
  x_{n+1} = \varphi^h_f (x_n) \approx x_n + h f(x_n),
  \label{eq:expEuler}
\end{equation}
where $h$ is the step size, which can be varied over subsequent steps. 
There exists a wide range of more complex numerical solvers, differing in their order of convergence, stability, and computational cost. One motivation for this work is to apply those methods to GP-based ODEs while also accounting for uncertainty. 
For details on dynamical systems and numerical integration see for example \cite{guckenheimerNonlinearOscillationsDynamical2013}.

A related class of models are discrete dynamical systems, which directly assume a model of the form 
\begin{equation}
  x_{n+1} = f(x_n),
\end{equation}
instead of discretizing a continuous model. 
Representing the discrete flow map with a GP is a more popular alternative to learning the vector-field, and the resulting model is called a \textit{state-space GP} [\cite{kamtheDataEfficientReinforcementLearning2017, buisson-fenetActivelyLearningGaussian2020, girardMultiplestepAheadPrediction2003,grootMultiplestepTimeSeries2011}]. 
This approach implicitly assumes a fixed step size $h$ between subsequent states $x_n$ and $x_{n+1}$, often the measurement interval of the available time series data. One computes a trajectory by iteratively applying $f$. 
Some approaches additionally use auto-regression, taking into account $l$ past states [\cite{grootMultiplestepTimeSeries2011, kocijanModellingControlDynamic2015, nghiemLinearizedGaussianProcesses2019}].

The challenge with uncertainty propagation for both continuous and discrete models is that mapping a random variable through a nonlinear function is generally intractable, even for the generally very tractable normal distribution.

In the context of dynamical system models, this problem is compounded. 
To obtain the distribution of the states $X_{n+1}, X_{n+2}$ and so forth in the trajectory of a state-space GP one has to repeatedly map the state $X_n$ through the distribution of nonlinear functions represented by the GP. 
For continuous models, the difficulty increases further. To compute subsequent states, we must compute a distribution of gradients $f(X)$ for the state $X$, which needs to be combined with the distribution of current state $X_n$ as seen in \eqref{eq:expEuler}. For higher-order methods, additional gradient distributions at intermediate states are required.

To our knowledge, this work is the first to consider approximate uncertainty propagation through numerical integrators for ODEs. However, there has been some work on uncertainty propagation in state-space GPs. 
Girard et al. published a series of results on approximating the output distribution when mapping a normal distribution through a GP, within the context of making iterative multiple-step-ahead predictions for discrete models.
In \cite{girardGaussianProcessPriors2002} the authors derive an approximation for the mean and variance of the output based on Taylor series approximations of the predicted mean and variance, using derivatives for the GP kernel function. A subsequent result in \cite{girardMultiplestepAheadPrediction2003} showed that when using the squared exponential kernel specifically, the mean and the variance of the output distribution can be obtained analytically. 
These results are called \textit{moment matching} in \cite{kamtheDataEfficientReinforcementLearning2017}.
Other work has applied sampling-based approaches to estimate uncertainty in state-space GPs [\cite{hewingSimulationTrajectoryPrediction2020}] and for continuous GP dynamics [\cite{hegdeVariationalMultipleShooting2022}].

The combination of ODEs and GPs has also been considered in the context of \emph{probabilistic numerics}. Well-known examples include \cite{schoberProbabilisticODESolvers2014} and the work mentioned in \cite{hennigProbabilisticNumericsComputation2022}. 
At a high level, ODE solvers under the probabilistic numerics umbrella consider a fully deterministic model and describe the continuous solution or trajectory as a Gaussian Process. 
Each discrete iteration step of the solver is then seen as a noisy observation, which makes it possible to quantify the uncertainty in the trajectory introduced by the integration algorithm. 
However, the setting in our work is orthogonal. The model itself is uncertain and we aim to include this uncertainty at least approximately in the trajectory prediction. 

We briefly summarize the setting in this paper.
We assume that the data-generating system has an unknown, but deterministic, dynamical system model. 
Given noisy observations, we express the different possible models and their likelihood 
as a distribution, represented for example by a finite GP conditioned on collected data. 

For simplicity, we completely omit all hyperparameter training and data conversion. Instead, we assume that model distribution is given, which allows us to focus on how to compute trajectory uncertainty from the model uncertainty.

Our main contributions are the following: 
We show that existing approaches to uncertainty propagation based on approximating the output distribution of the exact model assume the independence of the state and the model, and how this affects the predicted trajectory distribution.
We further propose an alternative approach based on a piecewise linear model approximation that can be solved exactly, resulting in what we call the PULL (Propagating Uncertainty through Local Linearization) class of solvers. 
We demonstrate the effectiveness of the PULL version of the explicit Euler and include discussions on its convergence and limitations.

\section{Review}
\subsection{Gaussian Processes}
We review briefly the basics of GPs [\cite{rasmussenGaussianProcessesMachine2006}].
We assume we have $N$ output observations $y^i = f(x^i) + \epsilon^i$ with Gaussian noise $ \epsilon^i \sim \mathcal{N}(0, \sigma^2_n)$ for known inputs $x^i$ to an unknown function $f: \R^m \rightarrow \R^d$. 
We express $f$ as a GP, i.e. as a distribution over functions.%
We specify a prior $\mathbb{P}(f)$ via a mean function $m(x)$, generally assumed to be zero, and a kernel function $k(x,x')$, which uniquely determines a Reproducing Kernel Hilbert Space containing all possible realizations of $f$.

For any finite subset of random variable outputs $f^i$ corresponding to known inputs $x^i$ the GP determines a joint Gaussian distribution, and by conditioning on the initial observations $\mathcal{D} = (x^i, y^i)$ we obtain the posterior distribution $\mathbb{P}(f|\mathcal{D})$, which allows us to predict the output $f(\hat{x})$ at a new input $\hat{x}$. 
Since we are considering a distribution of functions, we will obtain a distribution of outputs with the mean $\mu_f(\hat{x})$ and variance $\sigma^2_f(\hat{x})$ determined by 
\begin{subequations}
  \begin{align}
    \mu_f(\hat{x}) &= K_{\hat{x}, \mathbf{x}} (K_{\mathbf{x},\mathbf{x}}+\sigma^2_n I)^{-1} \mathbf{y} 
    \label{eq:GPmean}
    \\
    \sigma_f^2(\hat{x}) &= K_{\hat{x}, \hat{x}} - K_{\hat{x}, \mathbf{x}} (K_{\mathbf{x},\mathbf{x}}+\sigma^2_n I)^{-1} K_{\mathbf{x},\hat{x}}
    \label{eq:GPvar}
  \end{align}
\end{subequations}
where $\mathbf{x}=[x^1, \ldots, x^N]$, $\mathbf{y}=[y^1, \ldots, y^N]$, and $K_{x,x'} = [k(x^i, x'^j)]_{i,j}$ is the kernel or covariance matrix. 
The most common kernel is the squared exponential kernel
\begin{equation}
  k(x, x') = \exp\left( -\dfrac{1}{2} (x-x')^T W^{-1} (x-x')\right),
  \label{eq:sqexp-kernel}
\end{equation}
where $W = \text{diag}(w_1, \ldots, w_M)$ is the diagonal matrix of length scales.
We also use this kernel in our work, but our results hold for any kernel. 

In the context of dynamical systems the input and output dimensions are equal, hence $m=d$.
A number of vector-valued kernels exists \cite{alvarezKernelsVectorValuedFunctions2012}, but it is often assumed that each output can be treated independently with $d$ GPs $f:\R^m \rightarrow \R$.
For this work, we consider only one-dimensional dynamical systems and therefore one-dimensional GPs.

\subsection{Sampling Gaussian Processes}
GPs are distributions over function spaces which are generally infinite-dimensional and therefore inherently difficult to sample numerically. 
The standard option\,[\cite{wilsonEfficientlySamplingFunctions2020}] to compute the output of function realizations $f^*$ at specified input locations $x^*$ from the posterior distribution,  is to generate normally distributed random variables $\zeta \sim \mathcal{N}(0, I)$, and transform them according to the GP posterior such that
\begin{equation}
  f^* | \mathbf{y}, \mathbf{x}, x^* \ = m^* + K^{1/2}_{*,*} \zeta.
  \label{eq:gp-sample}
\end{equation}
Here, $(\cdot)^{1/2}$ indicates a matrix square root such as the Cholesky factor, $m^*$ is the GP mean $\mu_f(x^*)$ from \eqref{eq:GPmean} and $K_{*,*}$ is the covariance matrix for the sample input $x^*$ via \eqref{eq:GPvar}.
In essence, this is a \textit{grid-based} approach which, while numerically exact, is computationally costly as it scales cubically with the number of function values sampled.
To evaluate a function sample $f^*$ at arbitrary locations between grid points one can use standard interpolation methods. 

To mitigate the cubic scaling, alternative algorithms have been developed. 
Wilson et al. \cite{wilsonEfficientlySamplingFunctions2020} propose a combination of decoupled bases, specifically \textit{Fourier basis functions} \cite{rahimiRandomFeaturesLargeScale2007}, and kernel bases via a sparse GP approach. 
This has been used in other recent work \cite{ensingerStructurepreservingGaussianProcess2022,hegdeVariationalMultipleShooting2022} in the context of vector field models. 

An approach of this kind was also mentioned in \cite{hewingSimulationTrajectoryPrediction2020}, along with a \textit{memory-based} approach, which generates samples subsequently while conditioning each sample on at least some previous ones. 
It has also been used for uncertainty propagation in continuous dynamics [\cite{ridderbuschLearningODEModels2021}].

For this work we will use samples via \eqref{eq:gp-sample} as a source of ground truth for comparison, since it is the most accurate option to compare with.

\section{A Linear Perspective}
We begin with the simplest possible example, a prototypical linear model. We highlight the expected behaviour and introduce the approach by \cite{girardGaussianProcessPriors2002} based on approximating the output distribution of each subsequent state.
We then illustrate an issue with this approach for the prediction of trajectory uncertainty.

\subsection{A Linear Prototype}
\begin{figure}[t]
  \centering
  \begin{overpic}[width=0.48\textwidth]{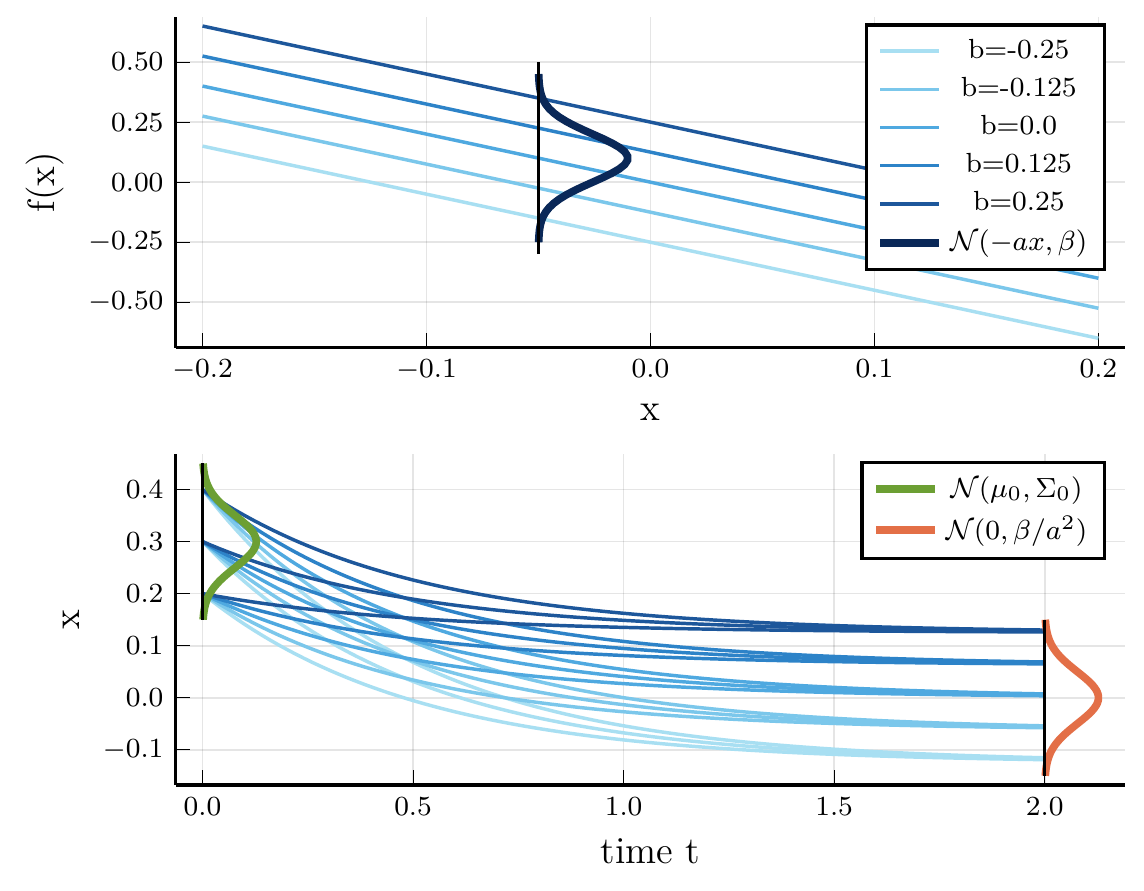}
    \put (1,75) {\textbf{a}}
    \put (1,38) {\textbf{b}}
  \end{overpic}
  \caption{\textbf{Linear Prototype.} \textbf{a:} The distribution of linear ODEs. \textbf{b:} Solving sampled linear ODEs for a distribution of initial values $\mathcal{N}(\mu_0, \Sigma_0)$ shows trajectories converging to the model parameter-dependent fixed point distribution. 
  }
  \label{fig:proto-lin-model}
\end{figure}

Consider a distribution of stable linear ODEs of the form 
\begin{equation}
  f(x) = -a x + B, \quad \text{with} ~ a\in \R^{+}, \,  B \sim \mathcal{N}(0, \beta),
  \label{eq:simple_linear_model}
\end{equation}
shown in Fig. \ref{fig:proto-lin-model}\text{a}, whose mean and variance are
\begin{equation}
  \mu(x) = -ax \quad \text{and} \quad \sigma^2(x) = \beta.
\end{equation}
The fixed point distribution of this distribution of ODEs is
\begin{equation}
  \hat{X} \sim \mathcal{N}(0, \beta/a^2),
  \label{eq:lin-model-fixed-point-distribution}
\end{equation}
which matches the behaviour shown in Fig. \ref{fig:proto-lin-model}b.
There, we sample realizations $b$ of $B$ to obtain deterministic linear ODEs, which we can solve for realizations $x_0$ of the initial value distribution $X_0 \sim \mathcal{N}(\mu_0, \Sigma_0)$.
As we will discuss in more detail in the following, the effect of the initial value is transient and trajectories converge to a fixed point only depending on the realization of the model variable $B$. 

\subsection{Approximating the Output Distribution}

Although the distribution of each subsequent state random variable $X_n$ is generally not normally distributed, the idea of \cite{girardMultiplestepAheadPrediction2003} is to compute its mean and variance and use a normal distribution with the same moments instead of the unknown general distribution.

Applying this approach to the example of the explicit Euler method \eqref{eq:expEuler}, a naive approach would be to apply it to the GP $f$ and then treat the sum of $X_n$ and $f(X_n)$ as independent. 
However, this leads to incorrect results, as they are clearly correlated. 

We can instead define a new function $g(x) = x + h f(x)$ and derive the equivalent expressions. Let the input $X_n$ be
\begin{equation}
  X_n \sim \mathcal{N}(\nu_n, \Sigma_n).
\end{equation}
The output of the GP for a specific sample $x^*$ of $X_n$ is then
\begin{equation}
  f(x^*) \sim \mathcal{N}(\mu(x^*), \sigma^2(x^*)).
\end{equation}
The distribution of ${X_{n+1} = g(X_n)}$ will generally not be normal, but we can at least compute its mean ${\nu_{n+1} = \mathbb{E}[X_{n+1}]}$ and variance ${\Sigma_{n+1}=\text{var}(X_{n+1})}$. 

From the law of iterated expectations it follows that the mean of the output distribution of one Euler step is given by
\begin{align}
  \nu_{n+1}(\nu_n, \Sigma_n) & = \mathbb{E}\left[X_n + h f(X_n)\right] \nonumber \\
    & = \nu_n + h\, \mathbb{E}[\mu(X_n)]
  \label{eq:output-approx-mean-iter}
\end{align}

The corresponding variance is given by 
\begin{align}
  \Sigma_{n+1}(\nu_n, \Sigma_n)  = & \text{var}\big(X_n + h f(X_n)\big) \nonumber \\ 
    = \, & \text{var}(X_n) + h^2 \text{var}\big(f(X_n)\big) \nonumber \\ 
    & + 2h \, \text{cov}\big(X_n, f(X_n)\big)
\end{align}
and from the law of total variance it follows that
\begin{align}
  & \text{var}(f(X_n)) \nonumber \\ 
  = ~& \mathbb{E}\left[\sigma^2(X_n) \right] + \mathbb{E}\left[ \mu(X_n)^2\right] - \mathbb{E}\left[ \mu(X_n)\right]^2.
\end{align}
This leaves the covariance between $X_n$ and $f(X_n)$. 
The law of total covariance states that
\begin{align}
  \cov\left(X_n, f(X_n)\right) =\, & \cov_{X_n}\left(\E_f[X_n|X_n], \E_f[f(X_n)|X_n]\right)  \nonumber \\ 
   & + \E_{X_n}[\cov_f(X_n, f(X_n) | X_n)] 
\end{align}

The first term can be written as 
\begin{align}
  & \cov_{X_n}\left(\E_f[X_n|X_n], \E_f[f(X_n)|X_n]\right) \nonumber \\
  =\, &\mathbb{E}\left[ X_n \mu(X_n) \right] - \nu_n \mathbb{E}\left[\mu(X_n) \right],
\end{align}
Adding only this term gives us 
\begin{equation}
  \begin{split}
    & \qquad \Sigma_{n+1}(\nu_n, \Sigma_n) = \Sigma_n \\ 
    &+ h^2\big(\mathbb{E}[\sigma^2(X_n)] + \mathbb{E}[\mu(X_n)^2] - \mathbb{E}[\mu(X_n)]^2 \big) \\
    &+ 2h\big(\mathbb{E}[X_n\mu(X_n)]
    - \nu_n \mathbb{E}[\mu(X_n)]\big).
  \end{split}
  \label{eq:non-hist-var}
\end{equation}
which matches the result in \cite{girardMultiplestepAheadPrediction2003} and \cite{grootMultiplestepTimeSeries2011}.
However, expression \eqref{eq:non-hist-var} is only correct if 
\begin{align}
  & \E_{X_n}[\cov_f(X_n, f(X_n) | X_n)] \nonumber \\ 
  =\, & \E_{X_n}\big[ \E_f[X_n f(X_n) | X_n] - X_n \mu(X_n) \big] \overset{!}{=} 0,
  \label{eq:cov-second-term}
\end{align}
which would imply that 
\begin{align}
  & \E_f[X_n f(X_n) | X_n] = X_n \E_f[f(X_n) | X_n] \nonumber \\
  =\, & X_n \mu(X_n). 
  \label{eq:wrong-pullout}
\end{align}
This is incorrect in general for trajectory predictions. The state
$X_n = X_n(f,X_0)$ is a function of the initial value and, critically, of the model $f$, which means that we cannot pull out $X_n$ in \eqref{eq:wrong-pullout}. 
Instead, to correctly compute the expected value over $f$ we must take into account how the distribution of models $f$ has affected $X_n$.

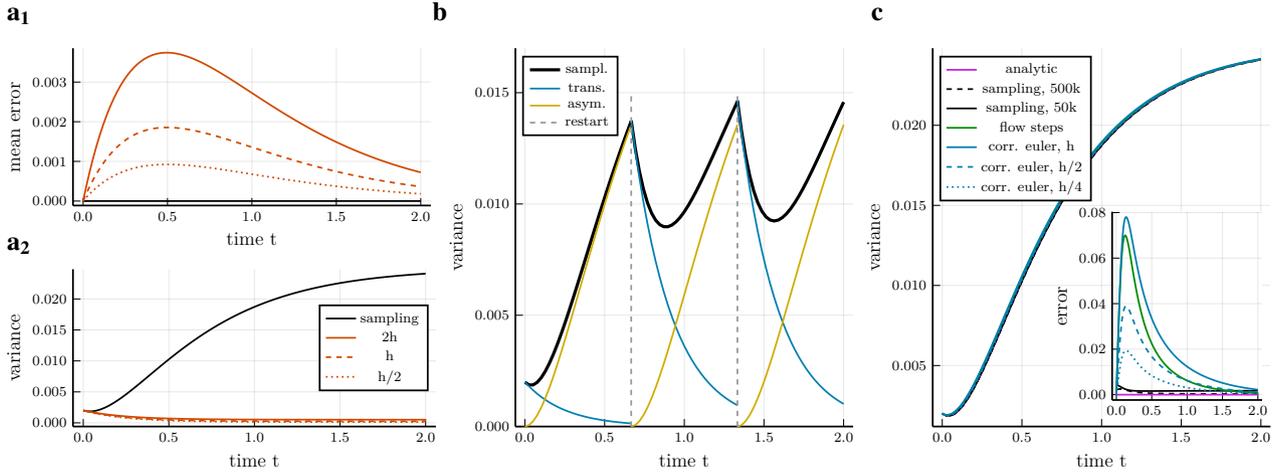
\begin{figure*}[t]
  \centering
  \begin{overpic}[width=0.99\textwidth]{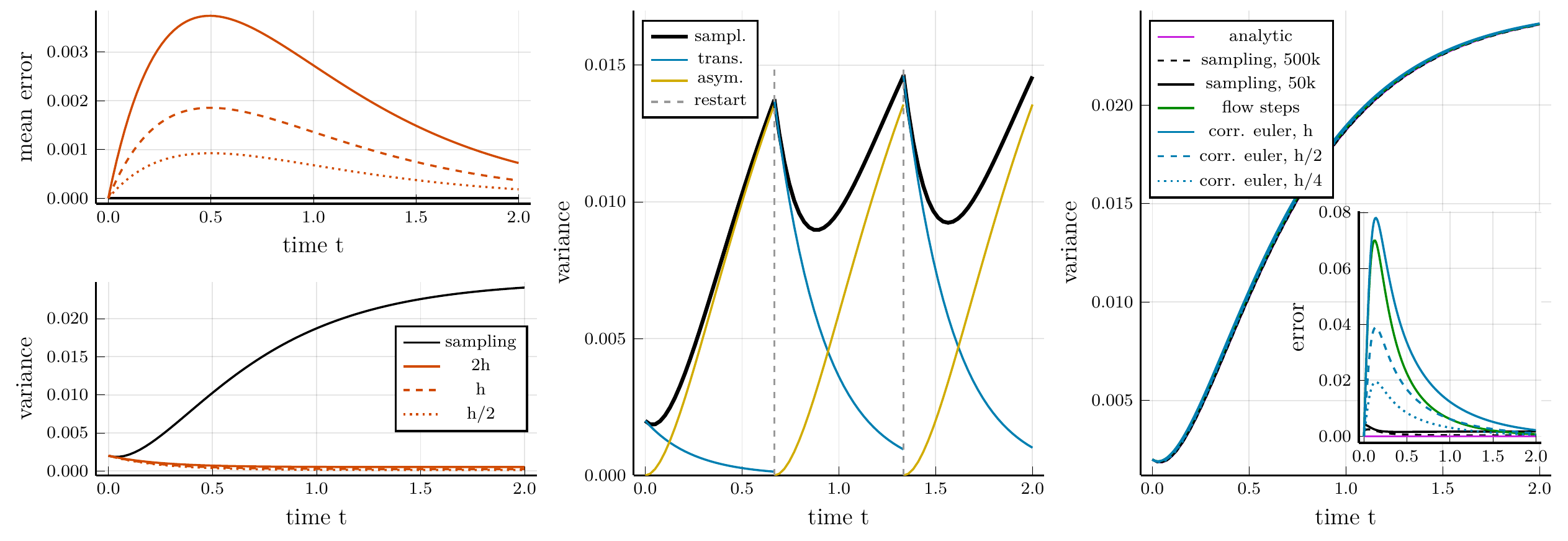}
    \put (1,36) {\textbf{a\textsubscript{1}}}
    \put (1,18) {\textbf{a\textsubscript{2}}}
    \put (34,36) {\textbf{b}}
    \put (68,36) {\textbf{c}}
  \end{overpic}
  \caption{\textbf{Computing Trajectory Distributions} \textbf{a:} With the independence assumption, decreasing the Euler step size decreases the mean error. However, the variance is vastly underestimated, and the error increases with smaller step sizes due to \eqref{eq:linear_bad_euler_var}. The reason is shown in \textbf{b}, where the black line represents the sampling result, with three restarts, each time again sampling from the final distribution of the previous segment. 
  It matches the sum of the analytic transient and asymptotic terms.   
  \textbf{c:} Iterating the flow and taking Euler steps with the correct covariance, we obtain results that match the analytic solution and sampling. Decreasing the step size correctly reduces the variance error compared to the analytic solution.
  }
  \label{fig:lin-model}
\end{figure*}

\subsection{Returning to the Linear Prototype}
We return now to the simple linear model \eqref{eq:simple_linear_model}.
The analytical solution for the flow of \eqref{eq:simple_linear_model} is
\begin{equation}
  \varphi^t (X_0) = \e^{-a t} X_0 + \dfrac{B}{a}(1 - \e^{-a t}).
  \label{eq:linear_flow}
\end{equation}
We assume that the initial value and the model parameter $B$ are independent, so $\text{cov}(X_0, B) = 0$. This results in the time-dependent mean and variance
\begin{subequations}
  \begin{align}
    \nu_t &= \mathbb{E}[\varphi^t (X_0)] = \e^{-a t} \mu_0 \\
    \Sigma_{t} &= \text{var}(\varphi^t (X_0)) \nonumber \\
    &= \e^{-2 a t} \Sigma_0 + \dfrac{\beta}{a^2}(1 - \e^{-a t})^2.
    \label{eq:cont_var}
  \end{align}
\end{subequations}
The expected values from the previous sections resolve to
\begin{subequations}
  \begin{align}
    &\mathbb{E}[\mu(X_n)] = -a \nu_n \\ 
    &\mathbb{E}[\sigma^2(X_n)] = \beta\\
    &\mathbb{E}[\mu(X_n)^2] = a^2(\Sigma_n + \nu_n^2)\\
    &\mathbb{E}[X_n \mu(X_n)] = -a(\Sigma_n + \nu_n^2),
  \end{align}
  \label{eq:lin-mod-exp-vals}
\end{subequations} 
which turns \eqref{eq:output-approx-mean-iter} and \eqref{eq:non-hist-var} into the iterations
\begin{subequations}
  \begin{align}
    \nu_{n+1} &= (1 - a h) \nu_n \\
    \Sigma_{n+1} & = \Sigma_n + h^2 \beta + h^2 a^2 \Sigma_n - 2 h a \Sigma_n.
    \label{eq:linear_bad_euler_var}
  \end{align}
\end{subequations}
 
Here we get the first indication that \eqref{eq:non-hist-var} is incorrect in this context. 
While the exact flow \eqref{eq:cont_var} results in the fixed point
\begin{equation}
  \hat{\Sigma}^{\text{exact}} = \dfrac{\beta}{a^2},
  \label{eq:exact-fixed-point}
\end{equation}
which matches the variance of \eqref{eq:lin-model-fixed-point-distribution}, the variance iteration map \eqref{eq:linear_bad_euler_var} results in 
\begin{equation}
  \hat{\Sigma}^{\text{euler}} = \dfrac{\beta}{a^2} \left(\dfrac{a h}{2 - a h}\right).
  \label{eq:euler-fixed-point} 
\end{equation}

The Euler steps converge to a fixed point distribution that depends on the step size $h$, which is also visible in Fig. \ref{fig:lin-model}a.
Even worse, the variance is vastly underestimated over the entire time span, and the error gets larger as the step size decreases. 
This is problematic -- the error of a method should generally not increase with smaller step sizes. 

Remarkably, we obtain a similar behaviour if we consider \eqref{eq:cont_var} as an iterable function of the state $X_i$ and step size $h$, 
\begin{equation}
  \Sigma^{h}_{X_n} = \e^{-2 a h} \Sigma_n + \dfrac{\beta}{a^2}(1 - \e^{-a h})^2.
  \label{eq:iter-flow-map}
\end{equation}
This map can be viewed as equivalent to repeatedly applying a state-space GP.
It has the fixed point 
\begin{equation}
  \hat{\Sigma}^{\text{iter. flow}} = \dfrac{\beta}{a^2} \tanh\left(\dfrac{a h}{2}\right)
  \label{eq:iterflow-fixed-point} 
\end{equation}
This is surprising, since the analytical flow map should have the semi-group property, such that $(\varphi^t \circ \varphi^s) (X_0) = \varphi^{t+s} X_0$. There should be no difference between applying the flow over multiple smaller intervals or over one large interval. 

To understand this behaviour, observe that \eqref{eq:linear_flow} is a sum of two terms.
For stable systems, the first term captures the \textit{transient} effect of the initial value, whereas the second term represents the \textit{asymptotic} behaviour determined by the model parameter $B$. 
If we apply the flow to the state $X_n$ instead of the initial value $X_0$, the state has already been affected by the model and decay of the initial transient has occurred.
The iterative schemes above are equivalent to starting over with an independent initial value after each time step, as can be seen in Fig.\,\ref{fig:lin-model}b.

The repeated transient explains the additional factor in $\hat{\Sigma}^{\text{iter. flow}}$, as smaller step sizes correspond to more restarts and initial transients in the same time period.  For $ah\rightarrow \infty$, we see either increasingly large steps or increasingly fast dynamics, which means that the transient behaviour fully decays in each step. 
The additional factor in $\hat{\Sigma}^{\text{euler}}$ similarly depends on $ah$. However, the expression breaks down as $ah\rightarrow 2$, the stability limit of the explicit Euler method.  

Therefore, the discrepancy between \eqref{eq:exact-fixed-point}, \eqref{eq:euler-fixed-point} and \eqref{eq:iterflow-fixed-point} arises from the covariance between the states and the model parameter, as mentioned at the end of the previous section.  
While it is valid to assume that $\text{cov}(X_0, B) = 0$, all subsequent states $X_i$ depend on the model, in this case on the random model parameter $B$, which means that they are not independent.
Instead, we find
\begin{align}
  \cov(X_n, B)
  = \,& \cov\left(\e^{-a n h}X_0 + \dfrac{B}{a}(1-\e^{-a n h}),\, B\right) \nonumber \\
  = \,&\dfrac{\beta}{a}(1-\e^{-a h n})
\end{align}
Adding this covariance to \eqref{eq:iter-flow-map}, we get
\begin{align}
  \Sigma^{h}_{X_n} =\, & \e^{-2 a h} \Sigma_n + \dfrac{\beta}{a^2}(1 - \e^{-a h})^2 \nonumber \\
  & + 2\dfrac{\beta}{a^2} (1 - \e^{-ah}) \e^{-ah}(1 - \e^{-ahn})
  \label{eq:linear_good_flow_map}
\end{align}

If $X_n$ is the result of subsequent Euler steps, 
\begin{align}
  X_{n} = & X_{n-1}+ h(-a X_{n-1} + B) \nonumber \\
  =\, & (1 - a h) X_{n-1} + h B
\end{align}
using a telescope sum we find $\cov(X_n, B)$ 
\begin{align}
  &\cov(X_n, B) = \cov\left((1 - a h) X_{n-1} + h B, \,B\right) \nonumber \\
  = \, & (1-ah) \cov(X_{n-1}, B) + h\, \cov(B, B) = \ldots \nonumber \\
  = \, & (1 - a h)^n \underbrace{\cov(X_0, B)}_{0} \nonumber \\ 
  &\, + h \sum_{i=0}^{n-1} (1-ah)^{n-1-i} \underbrace{\cov(B, B)}_{\beta}.
  \label{eq:lin-euler-cov}
\end{align}
We insert this into the corrected version of \eqref{eq:linear_bad_euler_var}
\begin{align}
  \Sigma_{n+1}=\, & (1-ah)^2 \Sigma_n + h^2 \beta \nonumber \\ 
    & + 2 h (1-ah) \cov(X_n, B).
    \label{eq:linear_good_euler_var}
\end{align}

We show the results of using \eqref{eq:linear_good_flow_map} and \eqref{eq:linear_good_euler_var} in Fig. \ref{fig:lin-model}c.
The variance from the iterative solutions matches the analytic solution and the variance of the sampled solutions (see Fig. \ref{fig:lin-model}c), and decreasing the step size correctly decreases the variance error. 
We also note that despite the simplicity of the model, we need to draw a substantial number of samples to match the analytical results.

This result is noteworthy beyond continuous dynamics. 
Applying a flow map like \eqref{eq:iter-flow-map} with some fixed step size $h$ iteratively without the additional covariance term is equivalent to repeatedly applying a learned state-space GP, as proposed in \cite{girardMultiplestepAheadPrediction2003} and applied in \cite{kamtheDataEfficientReinforcementLearning2017}. 
As this phenomenon is thus far unaddressed in previous work using moment matching for multiple-step-ahead predictions, the uncertainty has likely been underestimated. 

However, as the effect depends on the implicit step size $h$ in each step of the flow and might be affected by the inclusion of multiple previous points in the auto-regression context, further study is needed to understand the scope. 

\section{Local Linearization}

In this section, we address the general case distribution of models represented by a GP.  
Even discarding the correlation with past states, to use \eqref{eq:output-approx-mean-iter} and \eqref{eq:non-hist-var} to compute values for the mean $\nu_{n+1}(\nu_{n}, \Sigma_{n})$ and the variance $\Sigma_{n+1}(\nu_{n}, \Sigma_{n})$, we need expressions for the expected values $\mathbb{E}[\mu(X_n)]$, $\mathbb{E}[\sigma^2(X_n)]$, $\mathbb{E}[\mu(X_n)^2]$ and $\mathbb{E}[X_n \mu(X_n)]$. 
For the first three terms, we can find analytical expressions in \cite{girardGaussianProcessPriors2002} for the squared exponential kernel, and approximate expressions using linearization for all other kernels in \cite{girardMultiplestepAheadPrediction2003}. In \cite{grootMultiplestepTimeSeries2011} we also find the previously listed expressions for the squared exponential kernel and an expression for the term $\mathbb{E}[X_n \mu(X_n)]$.

However, we demonstrated in the previous section that we must include the correlation between the state and the model to accurately compute a trajectory, which means finding an approximation of \eqref{eq:cov-second-term}, which might be nearly intractable. 

Instead, we propose side-stepping the issue by approximating the entire GP model.
Specifically, we propose a piecewise linear approximation of the GP, defining a series of linear ODEs, which allows us to exactly propagate a normal distribution though each piece.
In other words, we approximate the model as a whole with a linearized model for which we can compute exact state distributions. 

We linearize around $\nu_i$ and define 
\begin{equation}
  f_i(x) = a_i x + B_i, \quad X(0) = X_i, ~ t \in [t_i, t_i+h),
  \label{eq:LL-ode}
\end{equation}
where 
\begin{align}
  a_i & = \mu_f'(\nu_i), \\
  B_i & \sim \mathcal{N}\left(\mu_f(\nu_i) - a_i \nu_i,\, \sigma^2_f(\nu_i) \right),
\end{align}
similar to \eqref{eq:simple_linear_model}.
While it would be more accurate to treat $a_i$ as a random variable, similar to the \textit{linGP} in \cite{nghiemLinearizedGaussianProcesses2019}, this would introduce additional complexity via the product of two random variables and additional covariances.

This approximation allows us to define the PULL (Propagation of Uncertainty through Local Linearization) class of ODE solvers for continuous GP models of ODEs.
However, it is currently not clear whether a similar approximation can be found for state-space GPs in order to incorporate the correlation with past states.

\begin{figure*}[t]
  \centering
  \begin{overpic}[width=0.99\textwidth]{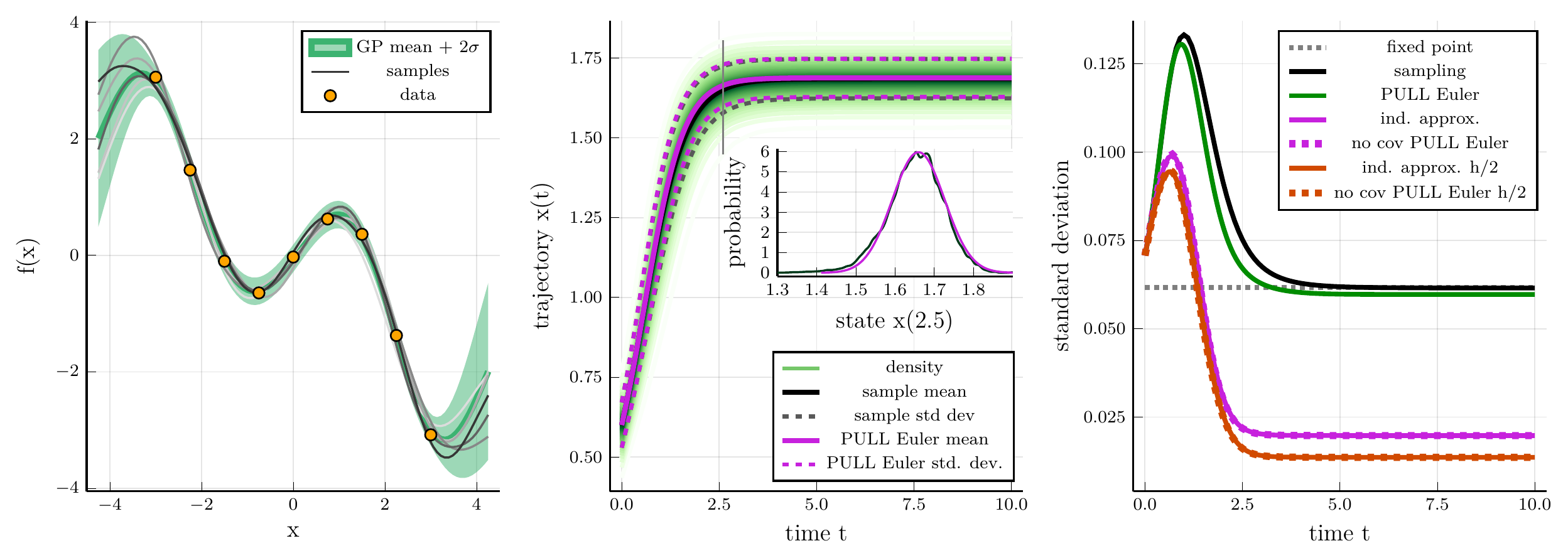}
    \put (0,31.5) {\large \textbf{a}}
    \put (34,31.5) {\large \textbf{b}}
    \put (68,31.5) {\large \textbf{c}}
  \end{overpic}
  \caption{\textbf{Nonlinear example.} \textbf{a:} The example function, including the noisy data points, the mean and the variance of the GP and a few samples from the GP distribution. 
  \textbf{b:} The distribution of the trajectories resulting from the GP samples, as well as their mean and standard deviation. 
  We compare with the results from the approximate local linearization-based solver. 
  \textbf{c:} The standard deviation resulting from sampling (ground truth), PULL Explicit Euler with full history, as well as the output approximation approach and PULL Euler without using any past points for two different step sizes each. 
  }
  \label{fig:non-linear-ex}
\end{figure*}

\subsection{Explicit Euler Method}
Taking Euler steps on each linear sections results in the iterative scheme
\begin{subequations}
  \begin{align}
    \nu_{n+1} =& \nu_n + h \mu_f(\nu_n) \\
    \Sigma_{n+1} =& (1+a_nh)^2 \Sigma_n + h^2 \sigma^2_f(\nu_n) \nonumber \\
    & + 2 h (1+a_nh)\, \cov(X_n, B_n).
    \label{eq:ll-euler-var}
  \end{align}
\end{subequations}
We use a telescope sum as in \eqref{eq:lin-euler-cov} for the covariance in \eqref{eq:ll-euler-var} and obtain 
\begin{equation}
  \cov(X_n, B_n) = h \sum_{i=0}^{n-1}\prod_{j=i+1}^{n-1}(1+a_jh) ~ \cov(\nu_i, \nu_n).
  \label{eq:ll-euler-cov}
\end{equation}
This expression requires storing all previous $a_i$, but as we store all trajectory states $X_i$, the storage requirements for the solver already scale linearly with the number of steps taken. 
The bigger challenge is computing the covariances $\cov(\nu_i, \nu_n)$ between the current state and all previous ones. 
This is done via \eqref{eq:GPvar} which means that the complexity scales quadratically with the number of steps already taken. 

To reduce this effort, there are two reasons to consider truncating the series in \eqref{eq:ll-euler-cov}. 
Firstly, we often operate within the basin of attraction of a fixed point. 
Then it will generally be the case that $a_i < 0$ and $\prod_{j=i+1}^{n-1}(1+a_jh) \rightarrow 0$ for $n\rightarrow \infty$. 
Secondly, when using a stationary kernel for the GP, the covariance is determined by the distance of the two points, where distant points are assumed to be nearly uncorrelated, which means $\cov(\nu_i, \nu_n) \approx 0$ for $i \ll n$. 

Both terms have complementary behaviour. Under stable dynamics, successive $x_n$ will be close and therefore correlated, but since we have $a_i < 0$ we can truncate based on the first term. 
Under unstable dynamics, we will have $a_i > 0$, but successive points will be further apart and the covariance term decreases faster. 

As the terms of the series are simple to compute, it is possible to implement adaptive truncation or include adaptive step sizes.
We also note that we have made no assumption about the underlying kernel beyond stationarity.

\section{Numerical Experiments}
In this section we illustrate the effectiveness of our PULL explicit Euler with a numerical example. \footnote{The code to reproduce all results and plots in this work is available under \texttt{github.com/<anonymous>}}
\subsection{A Nonlinear Example}
\label{sec:nonlin-example}
We consider the ODE
\begin{equation}
  \dot{x} = x \,cos(x), 
  \label{eq:cosx_model}
\end{equation}
and sample 9 data points in the interval $[-4, 4]$ and condition a GP with zero mean and the squared exponential kernel \eqref{eq:sqexp-kernel}.
Via \eqref{eq:gp-sample}, we generate $5000$ samples (as seen in Fig. \ref{fig:non-linear-ex}a) and integrate each one with initial values from $150$ samples of the input distribution $\mathcal{N}(0.6, 0.005)$, resulting in an empirical distribution of trajectories. We also apply the PULL explicit Euler to the same GP, starting from the same initial distribution, and find in Fig. \ref{fig:non-linear-ex}b that the results agree.

Further, we apply the extension \eqref{eq:non-hist-var} of the output approximation by \cite{girardMultiplestepAheadPrediction2003} and Groot et al. \cite{grootMultiplestepTimeSeries2011}, using their expressions for \eqref{eq:lin-mod-exp-vals} for the special case of the squared exponential kernel.
The results differ substantially from the sampling-based ground truth, and match our solver when not including any past states (Fig. \ref{fig:non-linear-ex}c). 
We again see that decreasing the step size increases the error compared to the sampling-based result.

\begin{figure*}[t]
  \centering
  \begin{overpic}[width=0.98\textwidth]{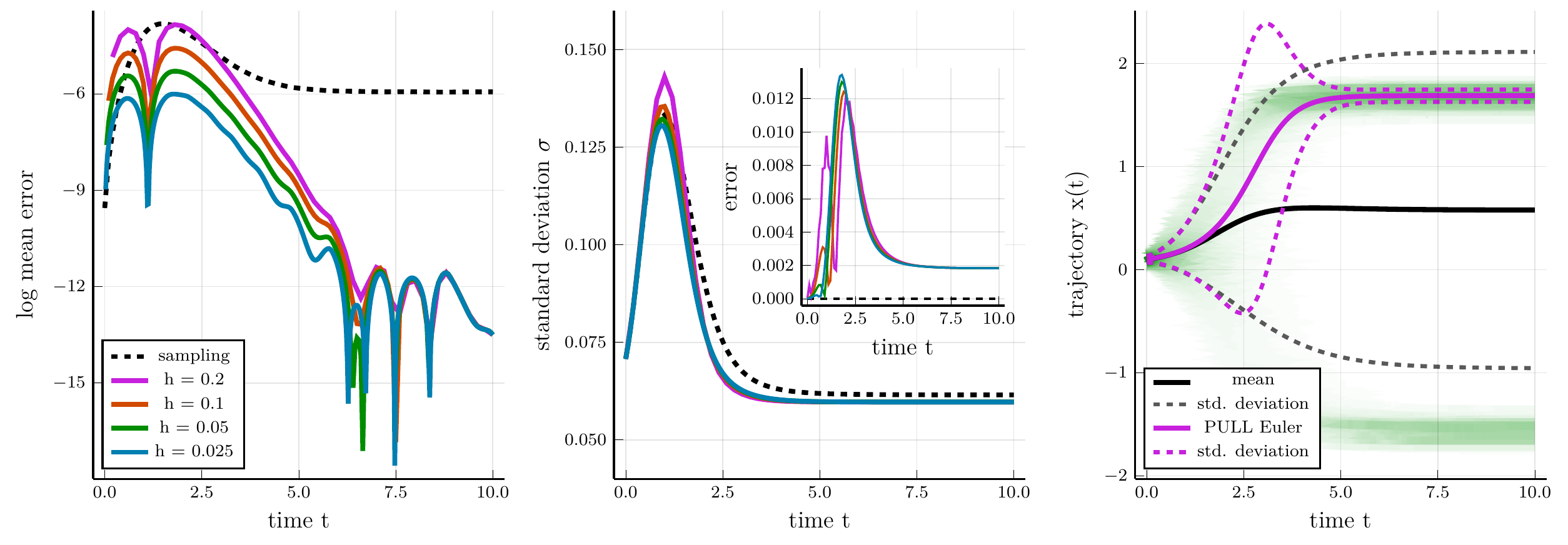}
    \put (1,33) {\large \textbf{a}}
    \put (33,33) {\large \textbf{b}}
    \put (67,33) {\large \textbf{c}}
  \end{overpic}
  \caption{\textbf{PULL Euler Convergence and Limitations} \textbf{a:} The logarithmic error of the solver compared to the solution of the ODE defined by the GP mean. Decreasing the step size also decreases the mean error. 
  \textbf{b:} The variance computed by our PULL solver is close to the sampling result. 
  \textbf{c:} Near an unstable fixed point, the sampled distribution of trajectories bifurcates towards the two nearby stable fixed points, while the linear approximation can only capture a single one. However, the increased transient variance compared to Fig. \ref{fig:non-linear-ex}b indicates hidden nonlinear dynamics. 
  }
  \label{fig:solver-results}
\end{figure*}

In Fig. \ref{fig:solver-results} we show the convergence of the PULL explicit Euler, which shows that compared to integrating the deterministic mean of the GP the error of our method correctly decreases and is lower than the sampling-based result. 
However, while the error of variance predictions compared to sampling-based result is acceptably low, it does not strictly decrease with decreasing step size, indicating some potential for improvement in our approximation of the model. 

Our PULL explicit Euler method returns accurate results as the estimated trajectory distribution converges towards a model-dependent fixed point distribution. 
The approximation slightly underestimates variance compared to the ground truth sampling, likely due to linearization error and can likely be improved upon.

\subsection{Limitations}

The linearization-based approach detailed in this work has substantially lower computational cost than a sampling-based approach, but has a fundamental limitation.
While linear models only have a single fixed point, nonlinear models introduce a wide array of additional behaviours, like additional fixed points and complex basins of attraction.  

In the model \eqref{eq:cosx_model} from section \ref{sec:nonlin-example}, there is an unstable fixed point in $0$ (see Fig. \ref{fig:non-linear-ex}a).
Hence, trajectories starting to the right of it will converge towards $\pi/4$ and ones starting on the left of $0$ will converge towards $-\pi/4$. 
Therefore, the distribution of trajectories starting from a distribution of initial values near $0$ will be bi-modal, as shown in Fig. \ref{fig:solver-results}c.
This behaviour cannot be captured by a linear approximation that assumes a uni-modal distribution for all states. 

Still, Fig. \ref{fig:solver-results}c highlights the usefulness of our efficient approximate solver. 
Solving the ODE defined by only the GP mean would completely miss the presence and effects of the unstable fixed point. 
Similarly, a UniversalODE [\cite{rackauckasUniversalDifferentialEquations2020}] using a Neural Network would likely also be oblivious to the presence of a nearby unstable fixed point. 

Our solver converges to the expected fixed point, as the initial value distribution has its mean to the right of $0$, but the standard deviation shows substantially increased initial uncertainty compared to the previous result from Fig. \ref{fig:non-linear-ex}b.
This provides a strong hint of non-linear behaviour, which can be investigated further with a more expensive, but more accurate, sampling-based method.

\section{Conclusion}
Accurately integrating dynamical systems expressed by GPs, either by numerically integrating a continuous vector field or subsequently applying a learned flow map, introduces a correlation between the GP distribution of functions and subsequent states. 
Using a linear prototype, we demonstrated that the approach in previous work incorrectly assumes independence between the distribution of the states and the GP, resulting in an underestimation of the trajectory variance.

We derived and illustrated the correct correlation for a simple linear model, and leveraged those findings to create a local linearization that can be combined with numerical integrator methods to create computationally efficient and accurate solvers, called PULL solvers. 

A possible application for these solvers is incorporating them into a complete pipeline from data to predictions.
We can create a likelihood-based cost function for training GP-based models with the predicted distribution of trajectories, as an alternative to multiple shooting methods [\cite{hegdeVariationalMultipleShooting2022}]. 
Once a model distribution is identified, we can make ahead-of-time predictions for model predictive control and take the prediction uncertainty into account when choosing a control strategy. 

Our results are very promising and highlight a fundamental consideration for predicting trajectories for GP-learned dynamical systems, suggesting several extensions. 
To develop methods with higher convergence order, one could improve the piecewise linear approximation or combine it with higher-order integration methods.

Extending this work to higher dimensions introduces a non-zero correlation between the components of each state $X_n$ and requires additional research. This correlation is not present in previous work due to using the independence assumption in \eqref{eq:cov-second-term}. 

Another interesting option is to combine our estimation of the model uncertainty with probabilistic numerics [\cite{hennigProbabilisticNumericsComputation2022}]. This would quantify the total uncertainty introduced by both the model, due to insufficient data, and the solver algorithm.

Lastly, a more in-depth study of the effects of the independence assumption in the context of state-space GPs is needed, as a smaller step size of the flow map often implies a higher measurement frequency in the time series data.
The resulting higher density of data already decreases the model uncertainty, and might partially mask the underestimation of the trajectory uncertainty .

\begin{acknowledgements} %
  S.R. acknowledges funding by the
  EPSRC Centre for Doctoral Training in Autonomous Intelligent Machines \& Systems EP/L015897/1.
\end{acknowledgements}

\newpage
\printbibliography
\end{document}